# Comparative Analysis of Advanced AI-based Object Detection Models for Pavement Marking Quality Assessment during Daytime

Gian Antariksa, Ph.D.
*PT Mitra Solusi Telematika*
Jakarta, Indonasia
gian.antariksa@gmail.com

Rohit Chakraborty
*Ingram School of Engineering*
*Texas State University*
San Marcos, Texas, United States
xuw12@txstate.edu

Shriyank Somvanshi
*Ingram School of Engineering*
*Texas State University*
San Marcos, Texas, United States
shriyank@txstate.edu

Subasish Das, Ph.D.
*Ingram School of Engineering*
*Texas State University*
San Marcos, Texas, United States
subasish@txstate.edu

Mohammad Jalayer, Ph.D.
*Department of Civil Engineering*
*Rowan University*
Glassboro, New Jersey, United States
jalayer@rowan.edu

Deep Rameshkumar Patel, Ph.D.
*Department of Civil Engineering*
*Rowan University*
Glassboro, New Jersey, United States
pateld80@rowan.edu

David Mills
*Department of Geography and Environmental Studies*
*Texas State University*
San Marcos, Texas, United States
davidmills@txstate.edu

*Abstract* — Visual object detection utilizing deep learning plays a vital role in computer vision and has extensive applications in transportation engineering. This paper focuses on detecting pavement marking quality during daytime using the You Only Look Once (YOLO) model, leveraging its advanced architectural features to enhance road safety through precise and real-time assessments. Utilizing image data from New Jersey, this study employed three YOLOv8 variants: YOLOv8m, YOLOv8n, and YOLOv8x. The models were evaluated based on their prediction accuracy for classifying pavement markings into good, moderate, and poor visibility categories. The results demonstrated that YOLOv8n provides the best balance between accuracy and computational efficiency, achieving the highest mean Average Precision (mAP) for objects with good visibility and demonstrating robust performance across various Intersections over Union (IoU) thresholds. This research enhances transportation safety by offering an automated and accurate method for evaluating the quality of pavement markings.

*Keywords* — Pavement Marking, YOLOv8, Object Detection, Transportation Safety

## I. INTRODUCTION

Longitudinal pavement markings, such as centerlines and edge lines, serve as essential guidance tools for defining the traveled path. By clearly indicating correct lane positions, these markings enhance driver safety, reducing the risk of head-on collisions, sideswipes with vehicles in similar lanes, and run-off-road (ROR) crashes [1]. Previous research on pavement markings has also demonstrated that higher retro reflectivity levels enhance driver visibility and increase the distance at which they can detect the markings [2]. Moreover, lane-keep assistance technology in autonomous vehicles also uses pavement markers. The quality of these pavement markers can also impact the safety of vehicles with lane-keep assistance technology [3], [4].

Additionally, the retroreflective pavement marking can significantly impact crash frequency [5], [6]. Better visibility of the markings and signs can increase the awareness of the drivers and a slight increase in driving speed [7]. Environmental conditions, application methods, and measurement inconsistencies can cause significant variability in pavement marking reflectivity [8]. This inconsistency poses challenges in accurately assessing the retroreflectivity of a roadway segment. In regions like New Jersey, reflective beads in the paint can wear out and be damaged by snowplows, necessitating regular restriping of pavement markings. These conditions necessitate the need for regularly inspecting the pavement marking quality. Therefore, it is crucial to regularly inspect and maintain pavement markings to ensure adequate visibility and roadway safety.

Previous studies have used deep learning techniques and computer vision to detect pavement marking quality [9], [10], [11]. Highlighting regions of interest (ROI) is fundamental in digital images, especially in pavement retro reflectivity quality classification. This procedure is essential for the detection and recognition of objects. Visual object detection deep learning models are categorized into one-stage and two-stage. Two-stage model types include Region-based Convolutional Neural Network (R-CNN) and Spatial Pyramid Pooling (SPPNet) in deep convolutional networks. These generate a pre-selected box called a region proposal (RP) that is likely to contain an object for detection [12]. These models then use convolutional neural networks to classify the identified samples.

Conversely, one-stage models extract visual features directly by skipping the RP step. Thus, these models determine the object's location and class. A few one-stage models include YOLO [13], Single Shot MultiBox Detector (SSD), and CenterNet [14], [15]. Previous research also stated the need for continuous checking of the pavement marking quality. This process, if conducted manually, may take longer time and can include human error. Therefore, this article focused on the detection of pavement marking quality using computer vision techniques.

The goal of the study is to develop and train a model for accurately detecting pavement marking quality based on their visibility to drivers during daytime. Utilizing image data sourced from New Jersey, this study employed three variants of the YOLOv8 model—YOLOv8m, YOLOv8n, and YOLOv8x—to assess pavement marking quality. The study leveraged transfer learning techniques to enhance model performance, based on pre-trained models to adapt to the specific task of pavement marking detection. By comparing the prediction accuracy of these three models, the research aims to identify the most effective model for real-time application.

## II. Literature review

Transportation safety is influenced by an array of factors, with roadway conditions and pavement marking quality playing crucial roles in driver behavior and crash rates. Poor pavement quality introduces distractions and compromises vehicle handling, particularly on high-traffic roads, highlighting the need for consistent maintenance to ensure safer driving environments [16]. Environmental conditions add complexity; for instance, rutting in pavements can lead to hydroplaning during rainfall, amplifying risks for vulnerable demographics, such as younger and older drivers, who may already face higher accident severity due to reduced control [17].

Traditional methods for assessing pavement marking quality rely on manual inspections, which require additional time and are subjected to human error. Therefore, an automated and efficient method to assess pavement marking conditions is required. This will ensure optimal road safety and maintenance technologies that can automate the pavement marking assessment including mobile imaging systems, Light Detection and Ranging (LiDAR), multispectral imaging, and machine learning (ML) and artificial intelligence (AI). Several case studies highlighted the effectiveness of automated pavement marking assessment and detection. A recent study explored vision-based pavement marking detection and condition assessment for Australian roads [18]. This research developed an automated framework for detecting and assessing pavement markings using image processing techniques. A study on automated methods for extracting pavement markings using mobile LiDAR data developed a novel process that utilizes information from the laser scanning point cloud, including red, green, and blue (RGB) color, intensity of laser pulse reflection, and differential intensity [19].

Recent advancements in deep learning and computer vision have enabled the development of robust object detection models essential for various transportation safety tasks. Deep learning models, for example, SSD, Faster R-CNN, and YOLO have demonstrated high accuracy and efficiency in detecting and classifying objects in images and videos [20]. The models can be divided into two-stage and one-stage detection models. Two-stage detectors, like Faster R-CNN, generate the region proposals at the beginning and classify them. At the same time, one-stage detectors, like YOLO, can directly predict class probabilities and bounding boxes. Models incorporating attention modules like the Convolutional Block Attention Module (CBAM) have shown an improved ability to focus on relevant image regions, enhancing detection accuracy in challenging conditions [21].

YOLO models have been applied in various domains, demonstrating their versatility and efficiency. For instance, in agriculture, YOLO models are used for fruit detection, enabling the classification of fruits based on ripeness [22]. In autonomous vehicles, YOLO models detect obstacles and pedestrians, enhancing navigation safety [23]. In another study, the YOLOv5 model was used for identifying near-miss events and non-compliance behavior to assess intersection safety [24]. These applications underscore the models' ability to handle diverse and complex detection tasks, making them suitable for assessing pavement marking quality. The speed of YOLO models is particularly advantageous in autonomous driving scenarios. For instance, the YOLO-Tomato model achieved detection times as low as 44 ms per image, demonstrating the potential for real-time object detection in dynamic environments [25]. The YOLO family of models has undergone significant evolution since its inception. YOLO models are useful in real-time prediction because of their accuracy and speed. The latest iteration, YOLOv8, introduces several architectural and functional improvements, such as the Cross-Stage Partial (CSP) module for efficient feature propagation, the Cross-stage 2-way feature fusion (C2f) module for superior visual feature extraction, and an anchor-free approach for dynamic alignment of classification and regression tasks [26]. These enhancements make YOLOv8 particularly effective for complex visual detection tasks. Traditional methods for assessing pavement marking quality involve manual inspections, which can require more time and are also subject to human error.

Despite significant advancements in automated pavement marking assessment using deep learning models, notable research gaps remain. Existing methods often face challenges related to class imbalance, where certain types of pavement markings are underrepresented in training datasets, leading to less accurate predictions. Additionally, many models attempt to achieve real-time processing capabilities, which are crucial for practical applications in transportation safety. The current study aims to fill these gaps using the YOLOv8 framework for pavement marking quality identification. The YOLOv8 model, with its architectural enhancements like the CSP and C2f modules and its anchor-free approach, promises more efficient feature propagation, superior visual feature extraction, and precise bounding box placement. By leveraging these capabilities, this study seeks to develop a robust and accurate model for real-time pavement marking assessment.

## III. METHODOLOGY

### A. Data Collection

This study aimed to assess road marking quality using retroreflective markings in New Jersey. The images are collected from the NJDOT Web Straight Line Diagrams (SLD) Data Browser. Images of 83 miles of road network have been collected comprised of 15,536 raw road images. These images include roadways that represent nine State Route Identifiers (SRIs). These are 00000007, 00000009, 00000023, 00000033_, 00000037, 00000046, 00000053, 00000072_, and 00000130_. Initially, the data was downloaded as photobook PDFs, and then the image data and roadway characteristics data, along with position, were extracted from the photobooks. The data collection and extraction process are illustrated in **Fig. 1**.

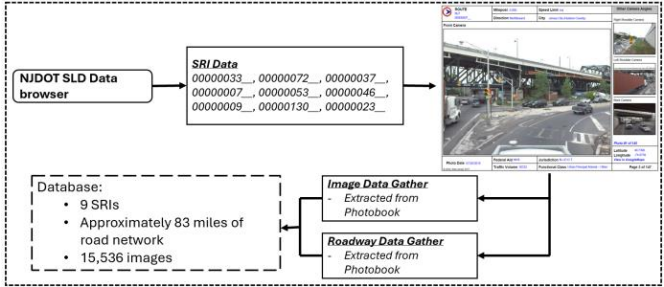

**Fig. 1** Data Collection and Extraction Process

In this study, the data collection process involved randomly selecting images and manually coding the visibility of pavement markings from the driver's perspective into three categories: good, moderate, and poor. This approach focused on three class variables related to marking quality. By employing YOLOv8, the current study processed an effective model for detecting the quality of pavement markings' visibility. Each dataset entry corresponds to an individual pavement marking quality based on visibility, allowing the exploration of unique circumstances across New Jersey roads. This study utilized 865 images out of a total of 15,536 available images for labeling and model training, which was then split into an 80% training set (692 images) and a 20% validation set (173 images). The decision to use a smaller subset was primarily driven by the time-intensive nature of manual annotation and the computational demands of training. A future step will involve labeling and using a larger portion of the 15,536 images to address this imbalance more effectively. This manual annotation ensured the visibility categories were accurately represented in the dataset. Once the training of the dataset was completed, the models were applied to the entire database. As illustrated in **Fig. 2**, the process ensures a comprehensive approach to model training and validation.

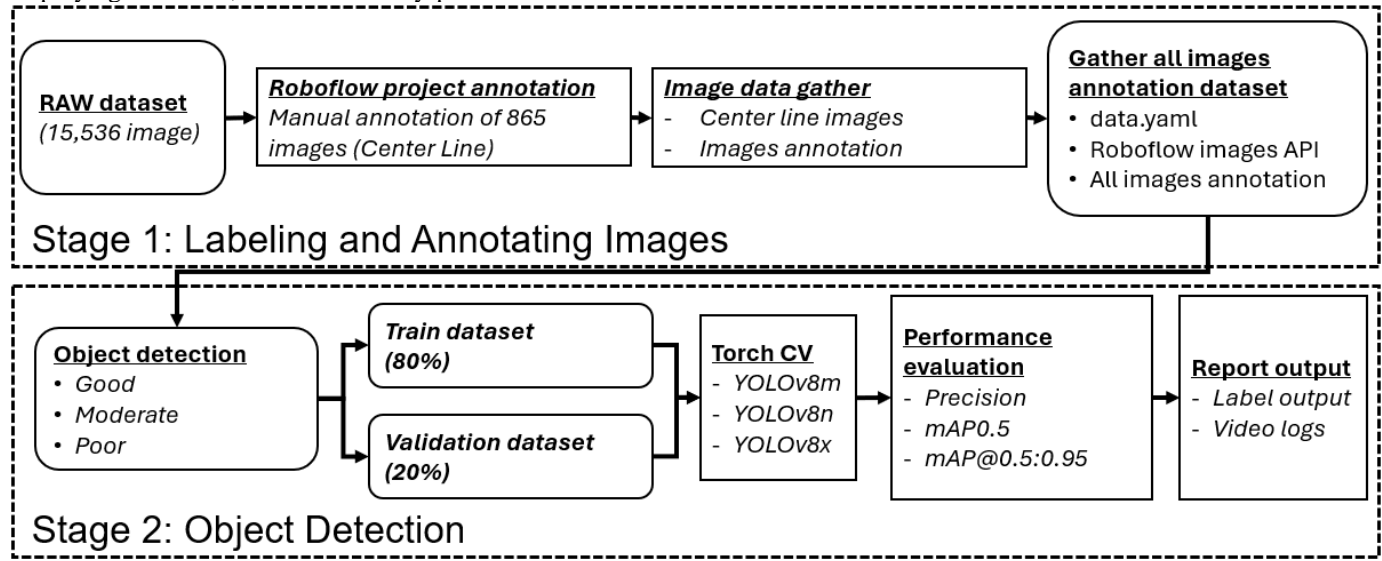

**Fig. 2** Experimental Process Flowchart

*B. You Only Look Once (YOLO)*

The YOLO algorithm is a deep learning network for real- In YOLO, the inputs are image pixel values, and it contains an output of predicted bounding boxes for objects that carry a confidence more significant than some defined value. It is a fast object detection algorithm using only one neural network to perform these tasks at high processing speeds [27]. The flow of the YOLO algorithm is shown in **Fig. 3**. The algorithm takes the frame as input and divides it into an NxN grid. Every cell in the grid must learn how to predict all those bounding boxes for objects within its domain. It also generates the class probabilities for these predicted bounding boxes simultaneously.

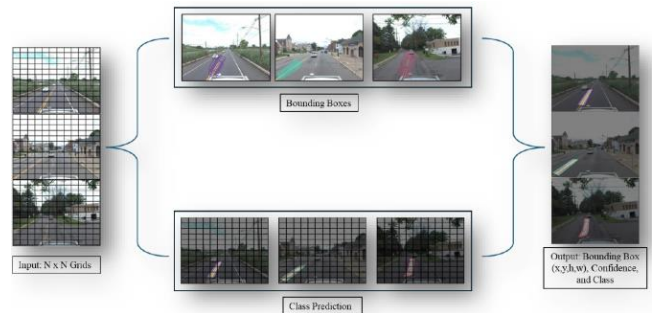

**Fig. 3** YOLO Algorithm Process Flowchart

*C. Model Theory*

The YOLOv8 model has an advanced object detection framework optimized for high precision and real-time processing, making it ideal for assessing pavement marking visibility. YOLOv8 integrates the CSP module and introduces the C2f module, significantly enhancing feature propagation and visual feature extraction. This dual-module approach allows the model to differentiate various pavement marking conditions effectively. A significant innovation in YOLOv8 is its anchor-free detection mechanism, which dynamically calculates an alignment metric from the classification score and the IoU between predicted and actual bounding boxes. This simplifies the model and enhances the precision of bounding box placements.

For classification tasks, YOLOv8 employs VFL, which addresses sample imbalance through asymmetric weighting as shown in Eq. (1). For regression tasks, DFL predicts multiple probability values for distances, improving localization accuracy as shown in Eq. (2).

$$VFL(p, q) = \begin{cases} -q(q\log(p) + (1-q)\log(1-p)), & q > 0 \\ \alpha p^{\gamma} \log(1-p), & q = 0 \end{cases} \quad (1)$$

$$DFL(p, q) = ((y_{i+1} - y)\log(S_i) + (y - y_i)\log(S_{i+1})) \quad (2)$$

*D. Limitation of YOLO*

Although YOLO provides real-time detection and good accuracy, it has several drawbacks. One challenge is detecting small objects in images, especially when they appear close to or behind larger objects. Another issue is that YOLO often requires a large labeled dataset, which may not be available for certain applications. Because YOLO relies on bounding boxes, it may also miss fine details of objects or markings when exact shapes or boundaries are important.

## IV. RESULTS AND DISCUSSIONS

*A. Model Evaluation and Validation*

The training and validation process involves an 80-20 split, where 80% of images are used to train and adjust the weights of the model and enhance its accuracy in identifying and classifying pavement markings. The remaining 20% is used for validation to evaluate the performance of the model on unseen data, ensuring reliable assessment in real-world scenarios. The primary metric for evaluating performance is the mAP across IoU thresholds from 0.50 to 0.95 (mAP50-95), providing a comprehensive view of accuracy. During training, the model's performance is validated at the end of each epoch using this metric, which allows for observing learning trends and detection accuracy. This epoch-wise evaluation aids in fine-tuning model parameters, identifying the optimal state, and avoiding overfitting. Consistent improvement in mAP scores indicates the benefit of continued training.

By applying these evaluation strategies, YOLOv8 ensures that the final model is robust, accurate, and reliable for practical applications in assessing the quality of pavement markings. This methodical training and validation approach supports enhancing road safety and infrastructure management through precise and dependable data analysis.

*B. Object Detection Results and Discussion*

YOLOv8 models are basic anchor-free. This study concentrated on using YOLOv8m, YOLOv8n, and YOLOv8x using three different weights. The obtained precision values from various epochs and models, were compared in similar experimental settings. Additionally, average inference time for real-time detection was utilized to assess the model's quality.

As previously mentioned, during the training process, the data samples were divided into two sets: 80% for training and 20% for validation. The performance of the model was estimated by calculating the loss separately for each set, resulting in an overall loss for the training data and a validation loss for the validation data. To evaluate the YOLOv8 training models based on the provided data, this study evaluated the performance metrics across different model variants (YOLOv8m, YOLOv8n, YOLOv8x) and training epochs (10, 25, 50, 100, 150). The metrics of interest are mAP0.5 and mAP@0.5:0.95, which provide insights into model accuracy at different IoU thresholds.

According to the model overview and performance metrics, this study used mAP0.5 and mAP@0.5:0.95, respectively. The mAP is measured by mAP0.5 at an IoU threshold of 0.5. This metric is easier for achieving high values, as it requires less overlap between the ground truth and predicted bounding boxes. Moreover, mAP@0.5:0.95 averages mAP calculated at IoU thresholds from 0.5 to 0.95. This is a more stringent and comprehensive metric, reflecting the model's performance across overlapping criteria. Table 1 indicates that with too few iterations, the model struggles to understand the quality of road marking reflectivity.

Table I Training Epoch Performance Metrics for YOLOv8 Model Variants

| Model | Weights | Epoch | mAP0.5 | mAP@0.5:0.95 |
|---|---|---|---|---|
| YOLOv8 | YOLOv8m | 10 | 0.285 | 0.119 |
| | | 25 | 0.373 | 0.169 |
| | | 50 | 0.489 | 0.227 |
| | | **100** | **0.494** | **0.247** |
| | | 150 | 0.468 | 0.232 |
| | YOLOv8n | 10 | 0.243 | 0.093 |
| | | 25 | 0.332 | 0.147 |
| | | 50 | 0.449 | 0.223 |
| | | 100 | 0.454 | 0.233 |
| | | **150** | **0.502** | **0.252** |
| | YOLOv8x | 10 | 0.288 | 0.124 |
| | | 25 | 0.397 | 0.181 |
| | | **50** | **0.493** | **0.245** |
| | | 100 | 0.469 | 0.240 |
| | | 150 | 0.473 | 0.238 |

Following the model output, YOLOv8m has mAP0.5 that increases notably from 0.285 at epoch 10 to a peak of 0.494 at epoch 100, suggesting significant improvements in identifying objects with at least 50% overlap with ground truths. On the model, YOLOv8n shows this model has a continuous improvement in both metrics up to 150 epoch, where it reaches its peak performance mAP0.5 at 0.502 and mAP@0.5:0.95 at 0.252, respectively, indicating steady learning and possible benefits from longer training durations without significant overfitting. In the last model, YOLOv8x starts with higher initial performance metrics than YOLOv8n but less than YOLOv8m. It peaks at epoch 50 for mAP0.5 (0.493) and mAP@0.5:0.95 (0.245). The performance then slightly declines, like YOLOv8m, suggesting potential overfitting or limitations in further learning past mid-training.

YOLOv8's anchor-free structure employs task alignment learning dynamic matching and incorporates DFL along with Complete IoU (CIoU) loss for regression branch function, ensuring high consistency between classification and regression tasks. During training data enhancement, disabling mosaic augmentation in the final ten epochs contributes to model convergence stability. As a general observation, on the early epochs, all models show substantial improvements in the early epochs (10 to 50), which is typical as the models learn basic features and patterns necessary for object detection. For the mid to late epochs, there are divergent trends post-50 epochs. At the same time, YOLOv8m and YOLOv8x show signs of peaking or slight declines, indicating potential overfitting, YOLOv8n shows continued improvements, suggesting different architectural or hyperparameter optimizations may be influencing robustness and learning capacity. In addition, due to extensive computational, model size and complexity, which offers crucial accuracy the optimal stopping point is needed for YOLOv8m and YOLOv8x and might benefit from stopping training around 100 epochs, whereas YOLOv8n could potentially explore even longer training periods. According to this scenario, YOLOv8n was chosen due to performance and highest accuracy, less computation resources due to time inference, it is proven by each class evaluation from Table 2. Based on the detailed analysis of each YOLOv8 model variant, YOLOv8m has precision peaks for object detection as "Good" at 0.775 on mAP0.5 and is significantly lower for "Moderate and "Poor" classes, with average inference time standing at 8.7 ms, which is efficient but not as fast as YOLOv8n indicating a balance between accuracy and speed. On the other hand, for YOLOv8x, it exhibits balanced precision across all classes, with the highest overall precision and mAP scores among the three variants. However, it has the slowest average inference time at 11.3 ms, which aligns with its enhanced accuracy and more complex computation. This led to the reason that YOLOv8n has the highest mAP0.5 for "Good" classes, about 0.823, and mAP@0.5:0.95 with 0.399, slightly surpassing YOLOv8m, making it exceptionally robust at detecting objects across various IoU threshold, with the average inference time about 3.4 ms which is the fastest among the variants, highlighting its suitability for scenarios where speed is most important.

## V. CONCLUSIONS

Despite significant advancements in automated pavement marking assessment using deep learning models, notable research gaps remain. Existing methods often face challenges related to class imbalance, where certain types of pavement markings are underrepresented in training datasets, leading to less accurate predictions. Additionally, many models struggle to achieve real-time processing capabilities, which are crucial for practical applications in transportation safety. These issues are prevalent in the current research of deep learning applications in pavement marking condition detection, as identified in recent studies. One of the primary challenges is the need for high-resolution images to capture the complex details of pavement markings accurately. Current methods often fall short in image quality, leading to inaccuracies in detection.

Table II Performance Comparison of YOLOv8 Model Variants in Object Detection

| Model | Weights | Class | Precision | mAP0.5 | mAP@0.5:0.95 | Average Inference time (ms) |
|---|---|---|---|---|---|---|
| YOLOv8 | YOLOv8m | Good | 0.775 | 0.790 | 0.388 | 8.7 |
| | | Moderate | 0.288 | 0.269 | 0.146 | |
| | | Poor | 0.481 | 0.474 | 0.235 | |
| | | All | 0.508 | 0.511 | 0.256 | |
| | YOLOv8n | Good | 0.647 | 0.823 | 0.399 | 3.4 |
| | | Moderate | 0.251 | 0.272 | 0.144 | |
| | | Poor | 0.400 | 0.420 | 0.224 | |
| | | All | 0.433 | 0.505 | 0.256 | |
| | YOLOv8x | Good | 0.685 | 0.788 | 0.382 | 11.3 |
| | | Moderate | 0.300 | 0.345 | 0.178 | |
| | | Poor | 0.387 | 0.421 | 0.215 | |
| | | All | 0.458 | 0.518 | 0.258 | |

This study addressed these gaps by employing the YOLOv8 pavement marking quality identification framework. The YOLOv8 model, with its architectural enhancements like the CSP and C2f modules and its anchor-free approach, promises more efficient feature propagation, superior visual feature extraction, and precise bounding box placement. By leveraging these capabilities, this study developed a robust and accurate model for real-time pavement marking assessment during daytime. The experimental results demonstrated that YOLOv8n stands out due to its optimal balance of performance, accuracy, and computational efficiency, making it suitable for scenarios where speed is crucial. YOLOv8n achieved the highest mAP0.5 for "Good" objects at 0.823 and a competitive mAP@0.5:0.95 of 0.399, combined with the fastest average inference time of 3.4 ms. This highlighted its suitability for real-time applications, contributing to enhanced road safety and maintenance.

Future research could explore further enhancing the efficacy of YOLOv8 models in pavement marking assessment. One potential area is the integration of multispectral imaging LiDAR as data improve the detection accuracy of various pavement marking types, especially in challenging conditions such as low-light or adverse weather. The ability to perform real-time, accurate assessments of pavement markings can lead to significant cost savings. By identifying and addressing deficiencies promptly, transportation agencies can avoid the higher costs associated with reactive maintenance and reduce the likelihood of severe roadway crashes caused by poor visibility. Developing a more extensive and diverse training dataset that addresses class imbalance issues can further refine the model's accuracy. Another promising direction is deploying these models in a real-world setting, such as integrating them into autonomous vehicles for continuous road monitoring. This would not only validate the model's real-time processing capabilities but also contribute to dynamic and adaptive road maintenance strategies, ensuring sustained road safety improvements. In addition, one of the primary challenges with the bounding box approach in lane marking quality detection is its limitation in capturing the detailed quality of pavement markings. Better algorithms that could potentially address these issues include the use of semantic segmentation models like U-Net, which can provide pixel-level accuracy and better handle class imbalance by focusing on the entire image rather than predefined regions. Additionally, transformer-based models such as Detection Transformer (DETR) could offer more precise and efficient feature extraction and classification capabilities, addressing the shortcomings of the traditional bounding box approaches in pavement marking quality detection. Although this study focused on daytime pavement marking detection, nighttime conditions can add challenges. However, it could not be implemented in this study as the considered dataset did not include any nighttime condition data. The limited visibility can reduce the contrast between markings and their surroundings, making it harder for the model to detect objects accurately. Headlight glare and uneven lighting also affect image quality. To address these challenges, future work could explore data augmentation with images that simulate low-light conditions, collect actual nighttime images for training, or incorporate specialized sensors (infrared or thermal cameras). Adapting the YOLO model to these varied lighting conditions may improve its reliability for round-the-clock road marking assessments. The current limitations of the present study can be explored in future studies.

ACKNOWLEDGMENT

This study was funded by the New Jersey Department of Transportation.